\documentclass[twocolumn, switch]{article} 

\usepackage{preprint}

\usepackage{amsmath, amsthm, amssymb, amsfonts}

\usepackage[numbers,square]{natbib}
\usepackage[utf8]{inputenc}	
\usepackage[T1]{fontenc}	
\usepackage{xcolor}		
\usepackage[colorlinks = true,
            linkcolor = purple,
            urlcolor  = blue,
            citecolor = cyan,
            anchorcolor = black]{hyperref}	
\usepackage{booktabs} 		
\usepackage{nicefrac}		
\usepackage{microtype}		
\usepackage{lineno}		
\usepackage{float}			

\usepackage{lipsum}		

\usepackage{multirow}

\usepackage{algorithmicx}
\usepackage{amsmath}
\usepackage{amssymb}
\usepackage{xcolor}
\usepackage[ruled,vlined]{algorithm2e}
\usepackage{colortbl}

\newtheorem{assumption}{Assumption}

\usepackage{newfloat}
\DeclareFloatingEnvironment[name={Supplementary Figure}]{suppfigure}
\usepackage{sidecap}
\sidecaptionvpos{figure}{c}

\usepackage{titlesec}
\titlespacing\section{0pt}{12pt plus 3pt minus 3pt}{1pt plus 1pt minus 1pt}
\titlespacing\subsection{0pt}{10pt plus 3pt minus 3pt}{1pt plus 1pt minus 1pt}
\titlespacing\subsubsection{0pt}{8pt plus 3pt minus 3pt}{1pt plus 1pt minus 1pt}

\usepackage{tikz,xcolor,hyperref}

\definecolor{lime}{HTML}{A6CE39}
\DeclareRobustCommand{\orcidicon}{
	\begin{tikzpicture}
	\draw[lime, fill=lime] (0,0) 
	circle [radius=0.16] 
	node[white] {{\fontfamily{qag}\selectfont \tiny ID}};
	\draw[white, fill=white] (-0.0625,0.095) 
	circle [radius=0.007];
	\end{tikzpicture}
	\hspace{-2mm}
}
\foreach \x in {A, ..., Z}{\expandafter\xdef\csname orcid\x\endcsname{\noexpand\href{https://orcid.org/\csname orcidauthor\x\endcsname}
			{\noexpand\orcidicon}}
}

\title{Fair Context Learning for Evidence-Balanced \\Test-Time Adaptation in Vision-Language Models}
\newcommand\titleheader{Fair Context Learning for Evidence-Balanced Test-Time Adaptation in Vision-Language Models}

\usepackage{xwatermark}

\usepackage{authblk}

\author{Sanggeon Yun}
\author{Ryozo Masukawa}
\author{SungHeon Jeong}
\author{Wenjun Huang}
\author{Hanning Chen}
\author{Mohsen Imani}

\affil{Department of Computer Science, University of California, Irvine\\\texttt{\{sanggeoy, rmasukaw, sungheoj, wenjunh3, hanningc, m.imani\}@uci.edu}}

\begin{document}

\twocolumn[ 
  \begin{@twocolumnfalse} 
  
\maketitle

\begin{abstract}
Vision--Language Models (VLMs) such as CLIP enable strong zero-shot recognition but suffer substantial degradation under distribution shifts. Test-Time Adaptation (TTA) aims to improve robustness using only unlabeled test samples, yet most prompt-based TTA methods rely on entropy minimization—an approach that can amplify spurious correlations and induce overconfident errors when classes share visual features. We propose \textbf{Fair Context Learning (FCL)}, an episodic TTA framework that avoids entropy minimization by explicitly addressing shared-evidence bias. Motivated by our \emph{additive evidence decomposition} assumption, FCL decouples adaptation into (i) augmentation-based exploration to identify plausible class candidates, and (ii) fairness-driven calibration that adapts text contexts to equalize sensitivity to common visual evidence. This fairness constraint mitigates partial feature obsession and enables effective calibration of text embeddings without relying on entropy reduction. Through extensive evaluation, we empirically validate our theoretical motivation and show that FCL achieves competitive adaptation performance relative to state-of-the-art TTA methods across diverse domain-shift and fine-grained benchmarks.

\end{abstract}  
\vspace{0.35cm}

  \end{@twocolumnfalse} 
] 



\section{Introduction}
\label{sec:intro}

Vision-Language Models (VLMs)~\cite{radford2021learning,jia2021scaling,abdul2023align,singh2022flava,zhai2023sigmoid,zhang2024vision} such as CLIP~\cite{radford2021learning} have demonstrated strong zero-shot generalization by aligning image embeddings with text embeddings derived from class prompts. Despite this capability, VLMs remain highly sensitive to domain shifts in background, texture, or visual context, which can significantly degrade recognition accuracy~\cite{mayilvahanan2023does,zhou2022learning}. A widely adopted strategy to improve robustness is Test-Time Adaptation (TTA), where the model adapts to each unlabeled test instance using only the information available at inference. TTA for VLMs commonly leverages multiple augmented views of a single test image (e.g., random crops, flips, or color jitters) to enhance stability and consistency during prediction.

Early TTA methods for VLMs primarily target the adaptation of the textual context embeddings. Test-Time Prompt Tuning (TPT)~\cite{shu2022test}, a pioneering approach, and its variants~\cite{wang2020tent,wang2022continual,yuan2023robust,zhang2022memo} optimize the textual context by minimizing prediction entropy across augmented views, thereby encouraging prediction consistency. However, entropy minimization alone is prone to pathological behavior under domain shifts. When several augmented views are biased toward incorrect but visually dominant cues, the entropy objective reinforces these biases, driving the model toward \emph{overconfident yet incorrect} predictions~\cite{guo2017calibration,mukhoti2020calibrating,eom2024adamer}.

\begin{figure}[t]
    \centering
    \includegraphics[width=0.5\textwidth]{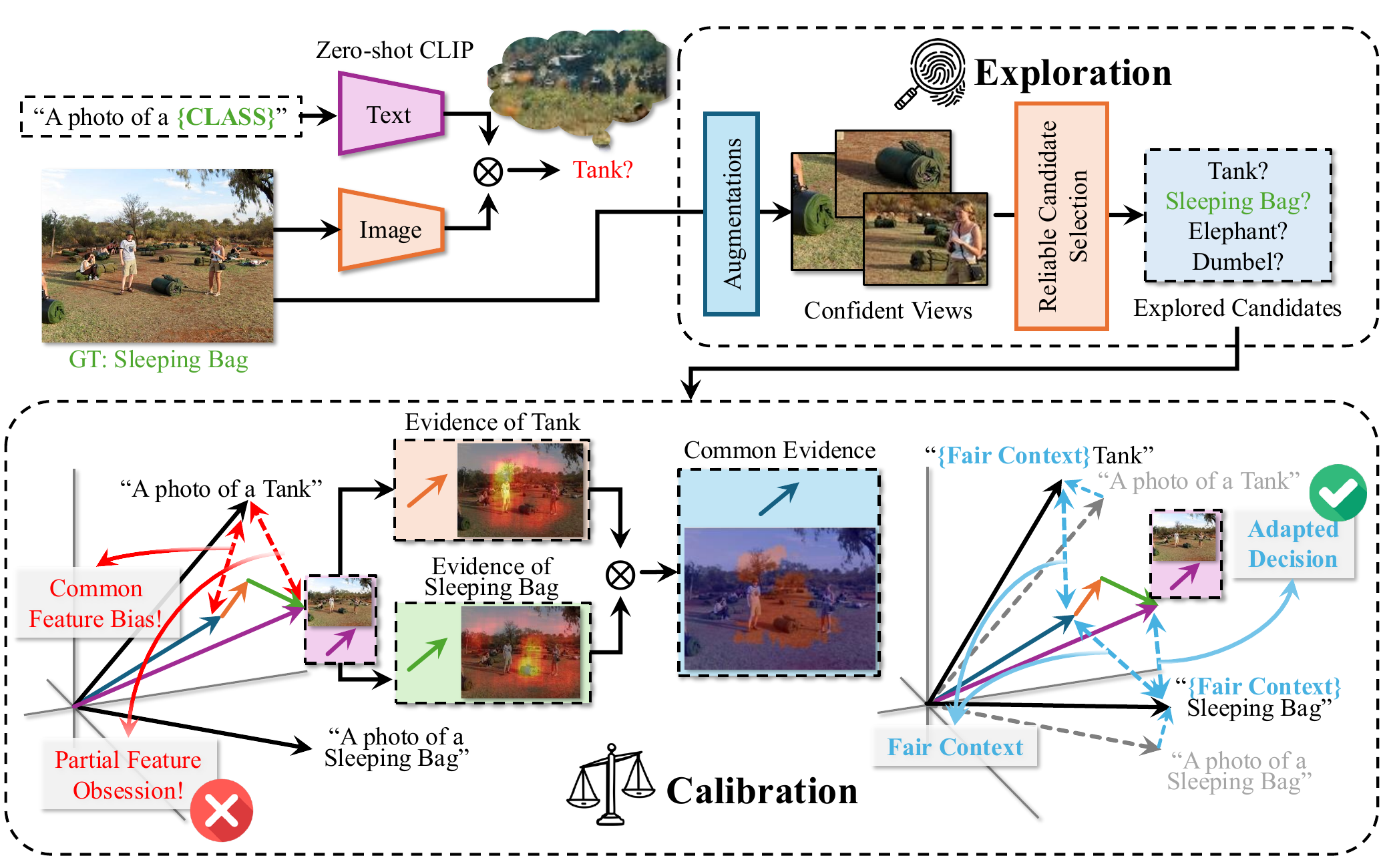}
    \caption{Conceptual overview of Fair Context Learning. Augmented views probe class-aligned evidence features. Our approach first explores candidate classes through confident view scoring and then calibrates the text embeddings in feature space using common-evidence maps, mitigating partial feature obsession.}
    \label{fig:FCLConcept_}
\end{figure}

To address the overconfidence issue, subsequent studies questioned the need for optimization-based adaptation and introduced simpler aggregation strategies~\cite{zanella2024test,farina2024frustratingly}, such as majority voting over confident views~\cite{farina2024frustratingly}. This approach improves robustness by suppressing inconsistent predictions and is theoretically justified by modeling view-level predictions as independent Bernoulli processes with fixed success probabilities. Under this assumption, majority voting serves as an ensemble estimator that asymptotically lowers the expected classification error relative to any single view. However, this formulation relies on a high-level independence assumption that neglects complex correlations among augmented views induced by shared visual features~\cite{liu2024understanding,tamkin2023feature}. As a result, it obscures the underlying mechanisms through which misclassifications persist across augmentations and fails to explain why both entropy-based and aggregation-based TTA approaches remain limited under domain shifts.

Our work departs from both entropy-minimization and independent-view aggregation paradigms. We instead conduct a concrete analysis of the mechanisms driving VLM misclassification under distribution shifts. We identify the key failure mode as \emph{partial feature obsession}---the model’s tendency to overemphasize common visual features shared across semantically related classes, biasing predictions toward incorrect categories while underutilizing class-unique evidence. This behavior emerges because augmented views primarily reweight the salience of shared evidence features rather than enhancing class-discriminative ones.
To illustrate, consider an image of a green ``\textit{Sleeping Bag}'' placed on a grassy field. Many augmented crops emphasize the green texture, elongated shape, and outdoor background. These features are also strongly present in images of the class ``\textit{Tank}'', which share similar visual contexts and occasionally include human figures. As a result, the model’s prediction across multiple augmented views can shift consistently toward ``\textit{Tank}'', even though distinctive evidence for ``\textit{Sleeping Bag}'' exists as conceptually depicted in \autoref{fig:FCLConcept_}.

Based on this observation, we propose a theoretical framework based on the \emph{additive evidence decomposition} assumption, which models an image embedding as a composition of (i) \emph{common evidence} shared among plausible classes and (ii) \emph{unique evidence} specific to each class. Under this formulation, augmentation acts as a mechanism that reweights these evidence components across views. When biased common evidence dominates, prediction confidence can concentrate on incorrect classes despite weak class-unique evidence, leading to the characteristic overconfidence observed in entropy-minimization-based TTA. Moreover, when most augmented views emphasize the same biased evidence, aggregation-based methods can also fail, as majority voting aggregation reinforces these biased views rather than filtering them out.

Building on this formulation, we propose \textbf{Fair Context Learning (FCL)}, a new episodic TTA framework that explicitly addresses the bias induced by shared visual evidence. FCL introduces a crucial \emph{decoupling} between two roles: (1) \emph{exploration}, where augmented views are used to identify a compact set of plausible candidate classes, and (2) \emph{calibration}, where the textual context is adapted to enforce \emph{fair sensitivity} across these candidates with respect to their shared evidence regions. Although exact evidence decomposition in the embedding space is intractable, we realize it through \emph{evidence map estimation}, which localizes image regions contributing to each candidate’s prediction and extracts pairwise common-evidence regions for calibration. \autoref{fig:FCLConcept_} illustrates both the decomposition of embedding space and the corresponding evidence maps used for context adaptation.

Our contributions are summarized as follows:
\begin{enumerate}
    \item We introduce a theoretical perspective grounded in the \textbf{additive evidence decomposition assumption}, explaining why entropy-based TTA and independence assumptions break under partial feature obsession, and empirically validate our theretical motivation.
    \item We propose \textbf{Fair Context Learning}, an episodic TTA framework that decouples adaptation into exploration and calibration: exploration identifies reliable candidate classes, and calibration adapts textual contexts by equalizing their sensitivity to shared visual evidence among those candidates.
    \item We show that FCL's non–entropy-based objective leads to consistent gains in robustness and generalization across a wide range of distribution-shift and fine-grained benchmarks.
\end{enumerate}

\section{Related Work}
\label{sec:related_work}

\paragraph{Vision--Language Models and Zero-Shot Learning.}
VLMs~\cite{radford2021learning,jia2021scaling,abdul2023align,singh2022flava,zhai2023sigmoid,zhang2024vision} such as CLIP~\cite{radford2021learning} learn joint image–text representations from web-scale data, enabling zero-shot transfer across diverse tasks. Classification is performed by comparing an image embedding to text embeddings derived from class prompts (e.g., ``a photo of a \{class\}''). Despite their success, VLMs remain sensitive to prompt design and distribution shifts in visual appearance, composition, and style.

\paragraph{Prompt Learning for VLMs.}
Prompt learning replaces hand-crafted templates with learnable textual contexts.  
Supervised approaches (e.g., CoOp, CoCoOp~\cite{zhou2022learning,zhou2022conditional}) fit prompts using labeled downstream data but often fail under domain shifts.  
Unsupervised and domain-aware methods~\cite{huang2022unsupervised,khattak2023maple} attempt to improve robustness, whereas we focus on \emph{unsupervised test-time} prompt adaptation, where only the test sample is accessible.

\paragraph{Test-Time Adaptation for VLMs.}
TTA adapts parameters during inference using unlabeled test inputs.  
Episodic TTA, which we adopt, adapts per instance.  
Entropy-minimization methods such as TPT~\cite{shu2022test} and its variants~\cite{yoon2024c,sheng2025r,imam2025test,sui2025just} update prompts using augmented views but often reinforce biased high-confidence errors.  
Beyond entropy, reinforcement-based updates~\cite{zhao2024testtime} and aggregation strategies ~\cite{zanella2024test, farina2024frustratingly} offer alternatives.  
Teacher-guided methods incur high overhead, while aggregation is efficient but cannot correct systematic misalignment between visual and textual evidence.

\section{Beyond Confidence Maximization}
\label{sec:motivation}

This section outlines the theoretical basis of our approach. We introduce the necessary background on VLMs and augmented-view TTA, then present a visual evidence decomposition model that explains how augmentations interact with image features and why confidence-based adaptation can fail under shared-feature bias. This analysis motivates a fairness-driven calibration mechanism, yielding our proposed FCL framework.

\subsection{Preliminaries}
\label{subsec:preliminaries_revised}

\paragraph{VLM Zero-Shot Classification.}
We consider a VLM comprising a visual encoder $f_v(\cdot)$ producing a normalized image embedding $z = f_v(x) \in \mathbb{R}^d$ ($||z||=1$) and a textual encoder $f_t(\cdot)$ producing a normalized text embedding $\tau_c(\delta) = f_t(c; \delta) \in \mathbb{R}^d$ ($||\tau_c(\delta)||=1$) for class $c$, parameterized by a learnable context $\delta$. The similarity score $s_c(x; \delta)$ uses the inner product, equivalent to cosine similarity for normalized embeddings, scaled by a temperature $\beta > 0$:
\begin{equation}
s_c(x; \delta) = \beta \langle z, \tau_c(\delta) \rangle.
\end{equation}
The probability $\pi(c \mid x, \delta)$ is obtained via softmax over the label space $\mathcal{Y}=\{1, \dots, C\}$.

\paragraph{Context Learning for TTA.}
While effective, VLM performance can degrade under domain shifts~\cite{radford2021learning}. TTA aims to mitigate this by adapting the model using only the unlabeled test image $x$. Many TTA methods employ \textbf{context learning}, optimizing the learnable context $\delta$ which parameterizes the textual encoder $f_t(c; \delta)$. Let $\tau_c(\delta)$ denote the context-adapted text embedding.

\paragraph{Augmented Views.}
TTA generates multiple \textbf{augmented views} $u_k = a_k(x)$ using stochastic transformations $a_k \sim \mathcal{A}$, where $\mathcal{A}$ defines a family of augmentation functions. Let $z_k = f_v(u_k)$ be the normalized embedding for view $k$. We denote scores, probabilities, and entropy on view $k$ as $s_{k,c}(\delta) = \beta \langle z_k, \tau_c(\delta) \rangle$, $\pi(c \mid x_k, \delta)$, and $H_k(\delta) = -\sum_{c} \pi(c \mid u_k, \delta) \log \pi(c \mid u_k, \delta)$, respectively. Let $y$ be the ground-truth label for $x$.

\subsection{Augmentation as Visual Evidence Probing}
\label{subsec:evidence_probing}

We posit that VLM errors often stem from sensitivity to shared visual features.

\begin{assumption}[Additive Evidence Decomposition]
\label{ass:decomp_revised}
The embedding $z_k$ of an augmented view $u_k$ can be approximately decomposed based on class-relevant visual evidence:
\begin{equation}\label{eq:additivedecomp}
z_k \approx z^{\mathrm{com}}_k + z^{\mathrm{uniq},y}_k + \sum_{j\neq y} z^{\mathrm{uniq},j}_k + \varepsilon_k,
\end{equation}
where $z^{\mathrm{com}}_k$ represents \emph{common evidence} shared among plausible classes, $z^{\mathrm{uniq},i}_k$ represents \emph{unique evidence} for class $i$, and $\varepsilon_k$ is residual noise. Augmentations $a_k$ primarily modulate the presence or salience (\emph{magnitudes}) of these components in the view $z_k$.
\end{assumption}

\paragraph{Modeling Score Contributions from Evidence Components.}
Under Assumption \ref{ass:decomp_revised}, we model how these decomposed visual evidence components influence the final score $s_{k,i}(\delta) = \beta \langle z_k, \tau_i(\delta) \rangle$. Applying the decomposition conceptually to the $z_k$, the score arises from the alignment of the text embedding $\tau_i(\delta)$ with each component:
\begin{equation}
\label{eq:score_decomp_revised}
\begin{split}
s_{k,i}(\delta) &\approx \beta \big( \langle z^{\mathrm{com}}_k, \tau_i(\delta) \rangle + \langle z^{\mathrm{uniq},i}_k, \tau_i(\delta) \rangle \\&+ \sum_{j\neq i}\langle z^{\mathrm{uniq},j}_k, \tau_i(\delta) \rangle + \langle \varepsilon_k, \tau_i(\delta) \rangle \big).
\end{split}
\end{equation}

This approximation helps illustrate how the score $s_{k,i}$ can be influenced by the presence and alignment of common features ($z^{\mathrm{com}}_k$), features unique to the target class ($z^{\mathrm{uniq},i}_k$), and potentially misleading features unique to other classes ($z^{\mathrm{uniq},j}_k$). The effectiveness of an augmentation $a_k$ in revealing class $i$ depends on how the resulting $z_k$ balances these alignments.

\paragraph{Low Entropy Signals Unique Evidence.}  
To understand how augmentations reveal class evidence, we analyze how prediction confidence relates to class-specific evidence.  
For view $k$ and class $i$, define the margin
\begin{equation}
m_k(i; \delta) = s_{k,i}(\delta) - \max_{j \neq i} s_{k,j}(\delta),
\end{equation}
which measures the score gap to the nearest competitor.  
A larger positive margin implies higher confidence for the true class $y$.  
From the softmax definition, one obtains the lower bound
\begin{equation}
\pi(c \mid u_k, \delta) \ge \frac{1}{1+(C-1)e^{-m_k(y;\delta)}},
\end{equation}
showing that large $m_k(y)$ guarantees high $\pi(y\mid u_k)$ and hence low entropy $H_k$.  
Thus, selecting low-entropy views is equivalent to selecting views with large positive margins.

Using the evidence decomposition (Assumption~\ref{ass:decomp_revised}), the margin expands as
\begin{equation}
\label{eq:margin_decomp}
\begin{split}
m_k(y;\delta) \approx \beta \Big( &\langle z^{\mathrm{uniq},y}_k, \tau_y(\delta) \rangle \\
 &- \max_{j\neq y} \big[ \underbrace{\langle z^{\mathrm{com}}_k, \tau_j(\delta) - \tau_y(\delta) \rangle}_{\text{Common Bias}} \\&+ \langle z^{\mathrm{uniq},j}_k, \tau_j(\delta) \rangle + R_{k,j}(\delta) \big] \Big).
\end{split}
\end{equation}
Here, the $R_{k,j}(\delta)$ represent residual terms arising from cross-unique alignments (e.g., $\langle z^{\mathrm{uniq},l}_k, \tau_j(\delta) \rangle$ for $l \neq y, j$) and noise ($\varepsilon_k$), which are assumed less dominant than the unique evidence and common bias terms. Therefore, low-entropy views correspond to those where the unique evidence term $\langle z_k^{\mathrm{uniq},y},\tau_y(\delta)\rangle$ dominates over common-bias and competitor contributions, confirming that entropy filtering naturally selects views containing strong class-specific evidence.

\subsection{Why Failures Occur: An Evidence-Based View}
\label{subsec:failure_modes}

\paragraph{Confidence Amplification.}  
When common evidence $z_k^{\mathrm{com}}$ in Eq.~\eqref{eq:margin_decomp} aligns more strongly with an incorrect class $j$ than with the true class $y$, the margin $m_k(y)$ becomes negative, leading to misclassification.  
Entropy-minimizing test-time adaptation then reinforces this bias by further amplifying the dominant but incorrect class—turning the context $\delta$ into a \emph{confidence amplifier} driven by partial evidence.

\paragraph{Limitations of Majority Voting.}  
While majority voting~\cite{farina2024frustratingly} reduces random view noise, it does not correct systematic text–image misalignment.  
If both $\tau_y$ and $\tau_j$ align with shared visual evidence, aggregated confident views remain biased.  
Voting thus improves robustness to noise but not fairness in representation.

\paragraph{Motivation for Fair Context Learning.}  
Under the additive evidence decomposition, failures arise when naive entropy minimization amplifies confidence on biased evidence, and majority voting, though effective for noisy view filtering, cannot resolve this bias.  
This motivates the need for \emph{fair context learning}—learning context as a calibration mechanism rather than a confidence amplifier.  
We therefore decouple test-time adaptation into two complementary stages: \emph{exploration}, where entropy-filtered augmentations identify reliable candidates, and \emph{calibration}, where fair context learning adjusts text embeddings to equalize shared evidence contributions and suppress bias amplification.

\section{Explore with Fair Context}
\label{sec:method}

\begin{figure*}[h]
    \centering
    \includegraphics[width=1.0\textwidth]{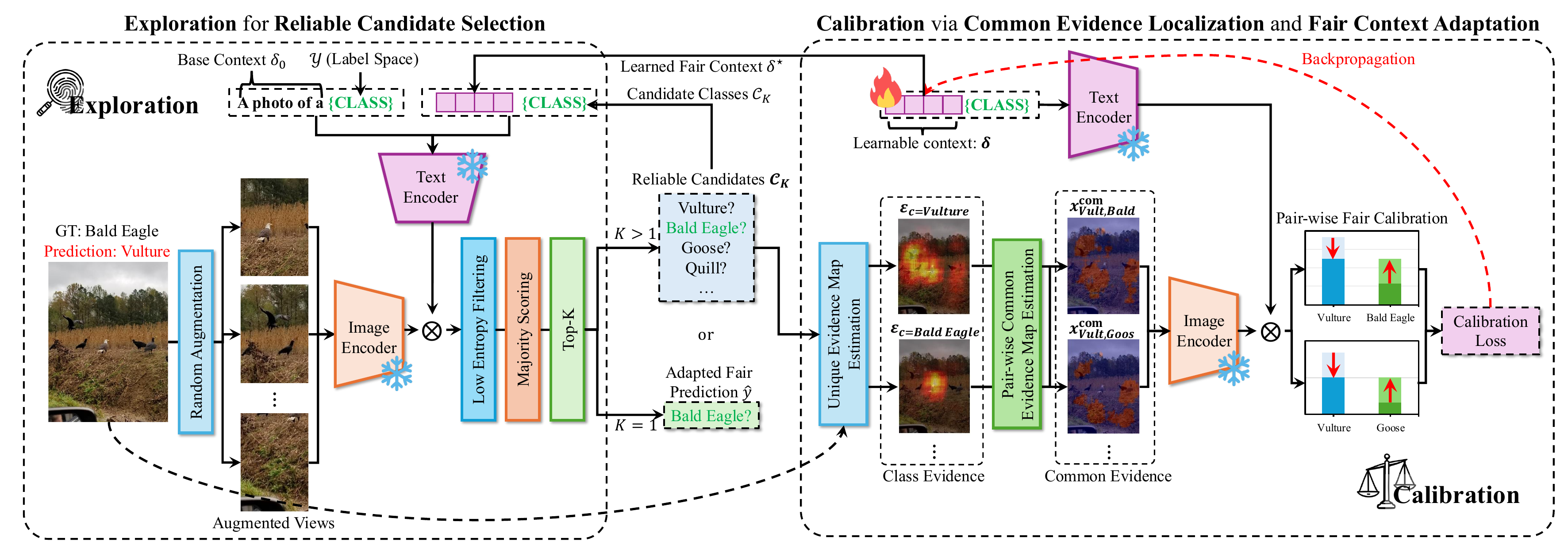}
    \caption{The pipeline of the proposed test-time adaptation framework via fair context learning.
Given a test image, the method first performs an exploration stage that identifies $K$ candidate classes $\mathcal{C}_K$ through augmented-view evaluation and low-entropy majority voting.
Next, in the calibration stage, the model learns a fair context by estimating pairwise common feature regions among candidate classes and optimizing their score distributions toward uniformity via backpropagation to update the soft prompt $\delta$.
Finally, the learned context $\delta^\star$ is applied for prediction by applying final exploration with the candidate classes $\mathcal{C}_K$.
}\label{fig:FCLTTA}
\end{figure*}

Building on our theoretical motivation, we introduce a framework that separates \emph{exploration} for reliable candidate discovery and \emph{calibration} for fair context learning as shown in \autoref{fig:FCLTTA}.  
The exploration stage identifies plausible class hypotheses using entropy-filtered augmentations with majority voting scoring, while the calibration stage refines text embeddings to balance shared evidence across them.

\paragraph{Exploration.}  
Given a test image $x$ and its augmented views $\{u_i\}_{i=1}^N$, the exploration stage selects top-$K$ reliable classes $\mathcal{C}_K \subset \mathcal{Y}$ by filtering low-entropy views—indicating distinctive, class-specific evidence—and aggregating predictions through majority voting to score each class to form $K$ reliable candidates:
\begin{equation}
\mathcal{C}_K = \Phi_K(x, \{u_i\}, \mathcal{Y}, \delta_0),
\end{equation}
where $\delta_0$ denotes the base context.

\paragraph{Calibration.}  
Given the candidate set $\mathcal{C}_K$, the calibration stage learns a context $\delta^\star$ such that the adapted text embeddings $\tau_c(\delta^\star)=f_t(c;\delta^\star)$ maintain \emph{equal sensitivity} to the shared visual features $z^{\text{com}}_{ij}$ between classes $i,j\in\mathcal{C}_K$:
\begin{equation}
\langle z^{\text{com}}_{ij}, \tau_i(\delta^\star)\rangle \approx \langle z^{\text{com}}_{ij}, \tau_j(\delta^\star)\rangle.
\end{equation}
This Fair Context Learning prevents overconfidence toward classes that dominate in common evidence.

\paragraph{Final Prediction.}  
The final prediction is obtained by reapplying the exploration process using the calibrated context:
\begin{equation}
\hat{y} = \Phi_{1}(x, \{u_i\}, \mathcal{C}_K, \delta^\star),
\end{equation}
clearly decoupling visual exploration for candidate identification from contextual calibration for bias-free inference.

\subsection{Exploration}
\label{subsec:candidate_selection_fcl}

\paragraph{View Generation and Filtering.}  
Following our theoretical finding that low-entropy predictions signal views rich in distinctive class evidence, we retain only the most informative augmentations.  
Given $N$ stochastic views $\{u_i\}_{i=1}^N$ of a test image $x$, candidate set $\mathcal{C}$, and context $\delta$, we compute image embeddings $z_i=f_v(u_i)$ and class scores $s_{i,c}(\delta)=\beta\langle z_i,\tau_c(\delta)\rangle$. The class posterior for class $c \in \mathcal{C}$ is defined as the softmax over all classes in $\mathcal{C}$:
$\pi_{\mathcal{C}}(c \mid u_i, \delta) = \frac{\exp(s_{i,c}(\delta))}{\sum_{j \in \mathcal{C}} \exp(s_{i,j}(\delta))}$. The entropy of the prediction on view $u_i$ is then:
\begin{equation}
H_i(\delta)=-\!\sum_{c\in\mathcal{C}}\pi_{\mathcal{C}}(c\mid u_i,\delta)\log \pi_{\mathcal{C}}(c\mid u_i,\delta).
\end{equation}
We retain the $\rho$ fraction of lowest-entropy views, indexed by $\Omega_\rho$.

\paragraph{Candidate Scoring and Identification.}  
To ensure stability against augmentation noise, class confidence is computed via majority voting. 
Each retained view casts one vote for its most probable class:
\begin{equation}
\hat{P}(c)=\frac{1}{|\Omega_\rho|}\sum_{i\in\Omega_\rho}\mathbf{1}\!\left[c=\arg\max_{j\in\mathcal{C}}\pi_{\mathcal{C}}(j\mid u_i,\delta)\right].
\end{equation}
The top-$K$ classes with the highest vote counts form the candidate set $\mathcal{C}_K = \Phi_K(x, \{u_i\}, \mathcal{C}, \delta)$, which is used for the calibration stage when $K > 1$, and directly yields the final prediction when $K = 1$.

\subsection{Calibration}

\subsubsection{Common Evidence Localization}
\label{subsec:common_evidence_fcl}

\paragraph{Proxy for Embedding Decomposition.}
Directly decomposing image embeddings $z_k$ into common and unique components (Assumption \ref{ass:decomp_revised}) is intractable in practice. Therefore, FCL employs a proxy by identifying regions in the \emph{input image space} that contribute significantly to the scores of multiple candidate classes inspired by prior work~\cite{Petsiuk2018rise}. These regions are assumed to contain the shared visual evidence relevant for calibration.

\paragraph{Class Evidence Map Estimation.}
For each candidate class $c \in \mathcal{C}_K$, we estimate a class evidence map $\mathcal{E}_c$ indicating the spatial importance of image regions for recognizing that class. This map helps localize potential visual evidence of the class. 
We employ a masking procedure where we sample $N_m$ masks $\{M_n\}_{n=1}^{N_m}$. 
Each mask $M_n$ is sampled at a randomly chosen spatial resolution from predefined grid sizes, masking a random fraction $\gamma$ of the image area. 
We create masked images $x^{(n)}$ by occluding (e.g., setting to black) the pixels of the original image $x$ within the region defined by $M_n$, while pixels outside $M_n$ remain unchanged. 
Our mask generation simplifies RISE~\cite{Petsiuk2018rise} by directly sampling multi-scale binary masks at target resolution, eliminating upsampling and interpolation overhead. 
We then measure the importance $\Delta\ell_c(M_n)$ of the occluded region for class $c$ as the increase in negative log-probability (using the base context $\delta_0$ and softmax restricted to $\mathcal{C}_K$) resulting from the masking:
\begin{equation}
\Delta\ell_c(M_n) = -\log \pi_{\mathcal{C}_K}(c \mid x^{(n)}, \delta_0) + \log \pi_{\mathcal{C}_K}(c \mid x, \delta_0).
\end{equation}
The underlying assumption is that if a masked region $M_n$ contains significant visual evidence supporting class $c$ (potentially unique features), its removal will decrease the model's confidence in class $c$, leading to a higher $\Delta\ell_c(M_n)$. The class evidence map $\mathcal{E}_c$ aggregates these spatially localized importance scores:
\begin{equation}
\mathcal{E}_c(u,v) = \frac{1}{N_m} \sum_{n=1}^{N_m} \Delta\ell_c(M_n) M_n(u,v),
\end{equation}
where $M_n(u,v)$ indicates whether pixel $(u,v)$ was part of the $n$-th mask. High values in $\mathcal{E}_c(u,v)$ thus suggest that pixel $(u,v)$ belongs to regions strongly associated with class $c$.

\paragraph{Common Evidence Map Estimation.}
Convert each $\mathcal{E}_c$ into a spatial probability map $\mathcal{S}_c$ via pixel-wise softmax. For each distinct pair of candidate classes $(i, j)$, the common-evidence map $\mathcal{Q}_{ij}$ identifies regions important to both:
\begin{equation}
\mathcal{Q}_{ij}(u,v) = \frac{\mathcal{S}_i(u,v) \mathcal{S}_j(u,v)}{\sum_{u',v'} \mathcal{S}_i(u',v') \mathcal{S}_j(u',v')} \in [0,1].
\end{equation}

The resulting $\mathcal{Q}_{ij}$ localizes regions jointly important to classes $i$ and $j$.  
Since direct decomposition of embeddings into $z^{\mathrm{com}}$ and $z^{\mathrm{uniq}}$ is intractable, we approximate the shared feature embedding by applying $\mathcal{Q}_{ij}$ pixel-wise to the image,  
$x^{\mathrm{com}}_{ij}=x\odot\mathcal{Q}_{ij}$,  
and compute $\tilde{z}^{\mathrm{com}}_{ij}=f_v(x^{\mathrm{com}}_{ij})\approx z^{\mathrm{com}}_{ij}$.  
This serves as a practical proxy approximation for the common evidence term in fair context calibration.

\subsubsection{Fair Context Adaptation}
\label{subsec:fair_adaptation_fcl}

We optimize the learnable context parameter $\delta$, starting from an initial state $\delta_0$. Define $\pi_{ij}^{(\delta)}(\cdot \mid \tilde{z}^{\mathrm{com}}_{ij})$ as the two-class softmax over $\{i,j\}$ using $\tilde{z}^{\mathrm{com}}_{ij}$ and the currently adapted embeddings $\tau_i(\delta), \tau_j(\delta)$. Let $\pi_{ij}^{(0)}(\cdot \mid z)$ be the corresponding base pairwise softmax using the full image embedding $z$ and the original base embeddings $\tau_i(\delta_0), \tau_j(\delta_0)$.

\paragraph{Calibration Loss ($\mathcal{L}_{\mathrm{cal}}$).}
To enforce fair calibration, we minimize the Jensen–Shannon (JS) divergence~\cite{lin2002divergence} between the adapted pairwise prediction on the approximated common feature $\tilde{z}^{\mathrm{com}}_{ij}$ and the uniform distribution $U=(\tfrac{1}{2},\tfrac{1}{2})$:
\begin{equation}
\mathcal{L}_{\mathrm{cal}}(\delta) = \frac{1}{|\mathcal{P}|} \sum_{(i,j)\in\mathcal{P}} 
{w_{ij}}\operatorname{JS}\!\left( \pi_{ij}^{(\delta)}(\cdot \mid \tilde{z}^{\mathrm{com}}_{ij}) \,\|\, U \right),
\end{equation}
where $\mathcal{P}$ is the set of unordered class pairs in $\mathcal{C}_K$, and 
$w_{ij}=1-\big|\pi_{\mathcal{C}_K}(i\mid x,\delta)-\pi_{\mathcal{C}_K}(j\mid x,\delta)\big|$
downweights pairs already balanced, emphasizing those exhibiting stronger bias.

\paragraph{Alignment Loss ($\mathcal{L}_{\mathrm{align}}$).}
To maintain semantic consistency, we regularize the adapted text features to stay close to their base embeddings:
\begin{equation}
\mathcal{L}_{\mathrm{align}}(\delta) = 1 - \frac{1}{K}\sum_{c \in \mathcal{C}_K} 
\cos\!\big(\tau_c(\delta), \tau_c(\delta_0)\big).
\end{equation}

\paragraph{Total Objective.}
The fair context $\delta$ is obtained by minimizing the combined loss:
\begin{equation}
\label{eq:total_loss_fcl}
\mathcal{L}_{\mathrm{total}}(\delta) 
= \lambda_{\mathrm{cal}}\mathcal{L}_{\mathrm{cal}}(\delta)
+ \lambda_{\mathrm{align}}\mathcal{L}_{\mathrm{align}}(\delta),
\end{equation}
and the optimized context parameter is denoted as $\delta^*$.

\subsection{Inference with Fair Context}
\label{subsec:inference_fcl}

Given the learned context $\delta^\star$ and the candidate set $\mathcal{C}_K=\Phi_K(x,\{u_i\},\mathcal{Y},\delta_0)$,  
we perform a final exploration using the adapted context to obtain  
\begin{equation}
\hat{y}=\Phi_{1}(x,\{u_i\},\mathcal{C}_K,\delta^\star).
\end{equation}
This produces the final consensus prediction from confident, calibrated augmentations under the learned fair context.

\section{Experiments}

\begin{figure*}[h]
    \centering
    \includegraphics[width=1.0\textwidth]{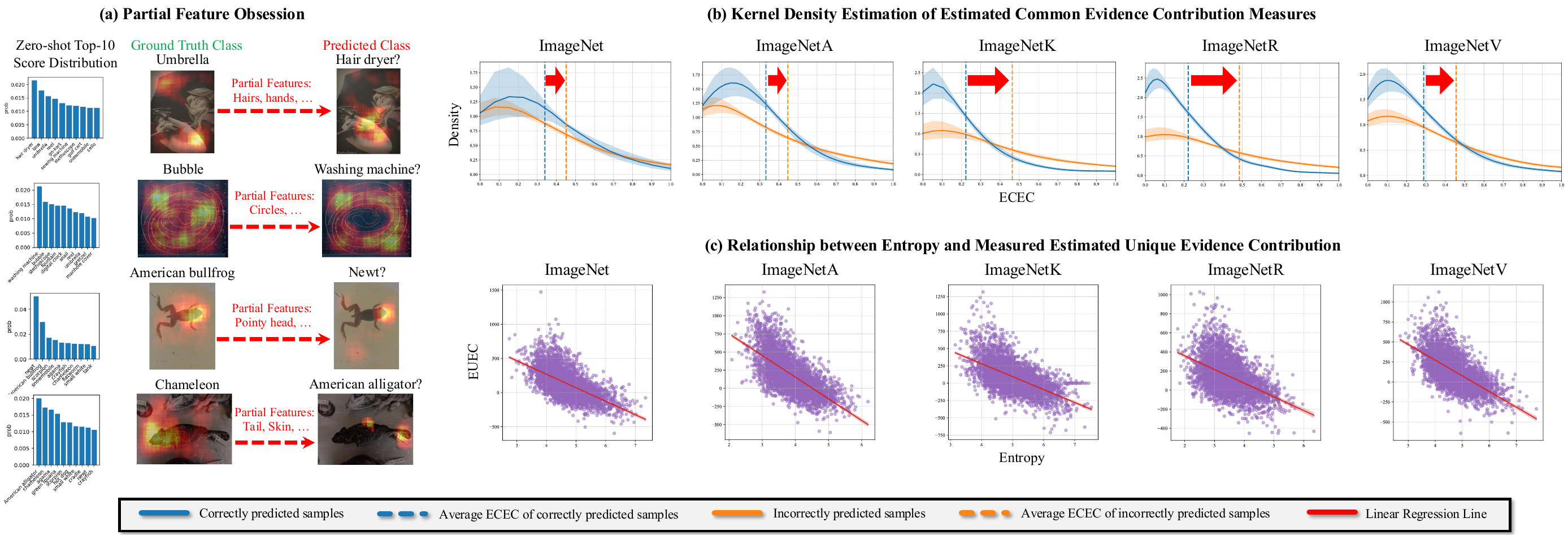}
    \caption{Empirical validation of our theory.  
(a) Partial Feature Obsession: score distributions and evidence maps show misclassifications driven by shared features.  
(b) Common Evidence Contribution: ECEC is consistently lower for correct samples across ImageNet variants.  
(c) Entropy–Uniqueness Relation: EUEC negatively correlates with entropy, indicating stronger unique evidence at low entropy.  
Shaded regions show 95\% CIs. Kernel density estimations are Gaussian with 400 bootstrap resamples.}
    \label{fig:observations}
\end{figure*}

\begin{table*}[t]
\centering
\caption{Top-1 accuracy (\%) of test-time adaptation methods on ImageNet and its distribution shift variants. 
The best result for each dataset is shown in \textbf{bold}, and the second best is \underline{underlined}. 
OOD Avg. is the mean over ImageNet-A, -K, -R, and -V.}
\label{tab:tta_imagenet_vitb16}
\resizebox{0.8\textwidth}{!}{
\begin{tabular}{l|ccccc|cc}
\toprule
Method & ImageNet & ImageNet-A & ImageNet-K & ImageNet-R & ImageNet-V & Avg. & OOD Avg. \\
\midrule
\rowcolor{gray!15}
CLIP~\cite{radford2021learning} & 66.72 & 47.80 & 46.15 & 73.99 & 60.84 & 59.10 & 57.20 \\
\midrule
TPT~\cite{shu2022test} & 68.94 & 54.75 & 47.92 & 77.13 & 63.38 & 62.42 & 60.80 \\
C-TPT~\cite{yoon2024c} & 68.52 & 51.35 & 47.48 & 75.87 & 62.60 & 61.16 & 59.33 \\
RLCF~\cite{zhao2024testtime} & 68.56 & 57.39 & 47.98 & 77.08 & 63.02 & 62.81 & 61.37 \\
MTA~\cite{zanella2024test} & 69.24 & 57.03 & 48.48 & 77.00 & 63.60 & 63.07 & 61.53 \\
ZERO~\cite{farina2024frustratingly} & 69.28 & \underline{59.77} & \underline{48.59} & 77.38 & 64.16 & \underline{63.84} & \underline{62.48} \\
TTL~\cite{imam2025test} & 69.28 & 59.00 & \textbf{48.75} & \underline{77.76} & \textbf{64.36} & 63.83 & 62.47 \\
TPS~\cite{sui2025just} & 68.83 & 58.19 & 48.24 & 76.98 & 63.68 & 63.18 & 61.77 \\
R-TPT~\cite{sheng2025r} & \underline{69.37} & 57.72 & 47.75 & 76.93 & 63.98 & 63.15 & 61.60 \\
\midrule
\rowcolor{blue!8}
Ours & \textbf{69.46} & \textbf{61.38} & 48.25 & \textbf{77.78} & \underline{64.23} & \textbf{64.22} & \textbf{62.91} \\
\bottomrule
\end{tabular}}
\end{table*}

\begin{table*}[t]
\centering
\small
\caption{Top-1 accuracy (\%) on fine-grained classification benchmarks. Best results are in \textbf{bold}, second-best are \underline{underlined}.}
\label{tab:tta_vitb16_fg}
\resizebox{1.0\textwidth}{!}{
\begin{tabular}{l|cccccccccc|c}
\toprule
Method & Aircraft & Caltech101 & Cars & DTD & EuroSAT & Flowers & Food101 & Pets & SUN397 & UCF101 & Avg. \\
\midrule
\rowcolor{gray!15}
CLIP~\cite{radford2021learning} & 23.91 & 94.00 & 65.51 & 44.39 & 42.22 & 67.40 & 83.64 & \textbf{88.25} & 62.56 & 65.24 & 63.71 \\
\midrule
TPT~\cite{shu2022test} & 23.55 & 94.24 & 66.24 & \textbf{47.04} & 43.04 & 68.62 & \underline{84.66} & 87.22 & 65.45 & \underline{68.41} & 64.85 \\
C-TPT~\cite{yoon2024c} & 23.94 & 93.71 & 65.71 & 45.27 & 42.47 & \textbf{69.43} & 83.17 & \underline{88.14} & 64.58 & 64.97 & 64.14 \\
RLCF~\cite{zhao2024testtime} & 22.26 & \textbf{94.44} & 66.41 & 46.04 & \underline{43.47} & 68.29 & 84.22 & 86.97 & 65.27 & 66.96 & 64.43 \\
MTA~\cite{zanella2024test} & \underline{24.84} & \underline{94.40} & \underline{67.49} & \underline{46.28} & 42.46 & 67.72 & 84.45 & 87.90 & 65.28 & 67.80 & \underline{64.86} \\
ZERO~\cite{farina2024frustratingly} & 24.81 & 94.00 & 67.03 & 45.39 & 37.28 & 67.44 & 83.78 & 87.33 & \underline{65.59} & 66.53 & 63.92 \\
TTL~\cite{imam2025test} & 24.75 & 93.75 & 66.25 & 45.86 & 39.02 & 66.38 & 83.99 & 87.22 & 65.07 & 67.30 & 63.96 \\
TPS~\cite{sui2025just} & 24.78 & 94.16 & 67.21 & 46.04 & 42.56 & 67.64 & 84.40 & 87.44 & 64.68 & 67.46 & 64.64 \\
R-TPT~\cite{sheng2025r} & 24.03 & 93.91 & 66.67 & 46.16 & 34.93 & \underline{69.02} & 84.28 & 86.73 & 65.50 & 67.35 & 63.86 \\
\midrule
\rowcolor{blue!8}
Ours & \textbf{34.17} & 91.76 & \textbf{68.58} & 45.45 & \textbf{43.69} & 67.36 & \textbf{86.57} & 86.54 & \textbf{65.71} & \textbf{68.52} & \textbf{65.83} \\
\bottomrule
\end{tabular}}
\end{table*}

\subsection{Experimental Setup}

We follow the standard experimental protocol of prior work~\cite{shu2022test,abdul2023align,sheng2025illusion}, evaluating on (i) natural distribution shifts and (ii) fine-grained classification, also referred to as \emph{cross-dataset generalization} in previous work.

\paragraph{Datasets.}
For natural distribution shifts, we use the ImageNet~\cite{deng2009imagenet} validation set together with ImageNet-A~\cite{hendrycks2021natural}, ImageNet-R~\cite{hendrycks2021many}, ImageNet-V2 (ImageNet-V)~\cite{recht2019imagenet}, and ImageNet-Sketch (ImageNet-K)~\cite{wang2019learning}, which are standard out-of-distribution (OOD) benchmarks for CLIP.  
For fine-grained and cross-dataset generalization, we use FGVC-Aircraft (Aircraft)~\cite{maji2013fine}, Caltech101 (Caltech101)~\cite{fei2004learning}, Stanford Cars (Cars)~\cite{krause20133d}, Describable Textures (DTD)~\cite{cimpoi2014describing}, EuroSAT~\cite{helber2019eurosat}, Oxford Flowers (Flowers)~\cite{nilsback2008automated}, Food101~\cite{bossard2014food}, Oxford Pets (Pets)~\cite{parkhi2012cats}, SUN397~\cite{xiao2010sun}, and UCF101~\cite{soomro2012ucf101}.  
For all datasets, we use the test splits defined in~\cite{zhou2022learning}, following the common protocol.

\paragraph{Implementation details.}
We build on the CLIP ViT-B/16 backbone~\cite{dosovitskiy2020image}.  
Images are resized to $224\times224$ and normalized using the CLIP mean and variance.  
For each test image, we use $N=64$ views: one original image and 63 stochastic augmentations obtained via random resized crops and horizontal flips.  
The text–image similarity temperature is set to $\beta=20$.  
For candidate selection, we retain the fraction $\rho=0.3$ of views with lowest predictive entropy and select the top-$K=10$ classes for adaptation.  
To estimate class evidence maps, we use $400$ random binary masks with multi-scale grids $\{7,9,11,13\}$.  
Prompt learning uses a 4-token context initialized from the template ``a photo of a'', optimized for $2$ gradient steps using AdamW~\cite{Loshchilov2017DecoupledWD} with a learning rate of $0.002$ and no weight decay.  
We set $\lambda_{\mathrm{cal}}=\lambda_{\mathrm{align}}=1.0$ and all other loss coefficients to zero.

\subsection{Empirical Validation of Theoretical Motivation}

\paragraph{Partial Feature Obsession.}
\autoref{fig:observations}.(a) illustrates the phenomenon of partial feature obsession in CLIP zero-shot inference.  
Each example shows the top-10 score distribution together with the estimated unique evidence maps of the ground-truth and misclassified classes.  
Although the ground-truth class features are present, CLIP often focuses on local shared patterns captured by $\tilde{z}^{\mathrm{com}}_{\hat{y}y}$—for instance, confusing “American Bullfrog’’ with “Newt’’ due to shared local patterns (e.g., textures or shapes).  
This confirms that the model’s prediction is driven by partial common evidence rather than full class-specific representations.

\paragraph{Common Evidence Contribution on Prediction.}
\autoref{fig:observations}.(b) quantifies this behavior using the \emph{Estimated Common Evidence Contribution} (ECEC), derived from our common-evidence formulation in Sec.~\ref{subsec:common_evidence_fcl}.  
For each predicted class $\hat{y}$ from zero-shot, we evaluate the average bias of its adapted text embedding toward shared evidence regions with other candidates $y \in \mathcal{C}_K \setminus \{\hat{y}\}$:
\begin{equation}
\mathrm{ECEC} =
\frac{\sum_{y \ne \hat{y}} \max\!\big(0,\, \langle \tilde{z}^{\mathrm{com}}_{\hat{y}y},\, \tau_{\hat{y}}(\delta_0) - \tau_y(\delta_0) \rangle \big)}
     {\sum_{y \ne \hat{y}} \max\!\big(\epsilon,\, \langle z,\, \tau_{\hat{y}}(\delta_0) - \tau_y(\delta_0) \rangle \big)}.
\end{equation}
A lower ECEC implies reduced bias toward shared features and greater reliance on unique class evidence.  
Across ImageNet and its variants, correctly predicted samples show consistently lower ECEC means and densities, while misclassified samples concentrate at higher values—empirically validating our theoretical claim that excessive alignment with common evidence drives systematic misclassification.

\paragraph{Validation of the Relationship between Low Entropy and Unique Evidence.}
\autoref{fig:observations}.(c) analyzes the relationship between prediction entropy and unique evidence strength across ImageNet variants.  
Each point corresponds to an augmented view $u_i$ and its \emph{Estimated Unique Evidence Contribution} (EUEC).  
We define EUEC as
\begin{equation}
\mathrm{EUEC} =
\frac{1}{K-1}\sum_{y \neq \hat{y}}
\langle \tilde{z}^{\mathrm{uniq}}_{\hat{y}y},\, \tau_{\hat{y}}(\delta_0) \rangle,
\end{equation}
where $\tilde{z}^{\mathrm{uniq}}_{\hat{y}y}=f_v(x^{\mathrm{uniq}}_{\hat{y}y})$ is obtained from the unique evidence region  
$x^{\mathrm{uniq}}_{\hat{y}y}=x\odot\mathcal{S}_{\hat{y}}\odot(1-\mathcal{Q}_{\hat{y}y})$,  
indicating areas important to $\hat{y}$ but not shared with class $y$.  
Across datasets, EUEC shows a strong negative correlation with entropy, confirming that low-entropy views emphasize discriminative, class-specific, unique evidence.  
This empirically supports our theoretical claim that entropy filtering isolates views enriched in unique features rather than merely confident ones.

\subsection{Evaluation Across Natural Shifts and Fine-Grained Domains}

We evaluate test-time adaptation performance on two widely studied regimes: (i) natural distribution shifts using the ImageNet and its variants, and (ii) fine-grained recognition benchmarks. Comparisons are made against zero-shot CLIP~\cite{radford2021learning} and eight \emph{reproducible} episodic TTA baselines for VLMs~\cite{shu2022test,yoon2024c,zhao2024testtime,zanella2024test,farina2024frustratingly,imam2025test,sui2025just,sheng2025r}; methods without publicly available code are excluded.%

\paragraph{Natural Distribution Shifts.}
\autoref{tab:tta_imagenet_vitb16} summarizes results on ImageNet and four commonly used OOD variants.
Zero-shot CLIP exhibits substantial degradation under these shifts, particularly on ImageNet-A and ImageNet-Sketch.
All TTA methods provide noticeable gains, confirming the benefit of episodic adaptation.
Among prior approaches, our method achieves the highest accuracy on ImageNet as well as its OOD variants—ImageNet-A and ImageNet-R—and obtains the best overall average performance (64.22\%).
The improvement on ImageNet-A is particularly significant, highlighting that our calibration-guided prompt refinement effectively resolves confusions induced by hard-to-recognize content.
The consistent boost across all four OOD test sets demonstrates that our approach generalizes beyond any single shift type.

\paragraph{Fine-Grained Classification.}
\autoref{tab:tta_vitb16_fg} shows that performance varies widely across fine-grained datasets.  
Aircraft is the hardest case—CLIP zero-shot is only 23.91\%, the lowest among all datasets—and our method improves over the next-best TTA approach by more than +9 pp, contributing strongly to the best overall average (65.83\%), and on Food101 and UCF101 we obtain the strongest performance among all methods.
\
In contrast, Caltech101 and Pets exhibit strong CLIP zero-shot performance (94.00\% and 88.25\%), indicating that their text–image alignment is already well calibrated.  
On these near-saturated regimes, our method yields lower accuracy, suggesting that fairness-based calibration may introduce noise when the baseline is already well aligned. 
This highlights the need for more conservative calibration strength in such cases.

\subsection{Ablation Study}

\begin{figure}[t]
    \centering
    \includegraphics[width=0.48\textwidth]{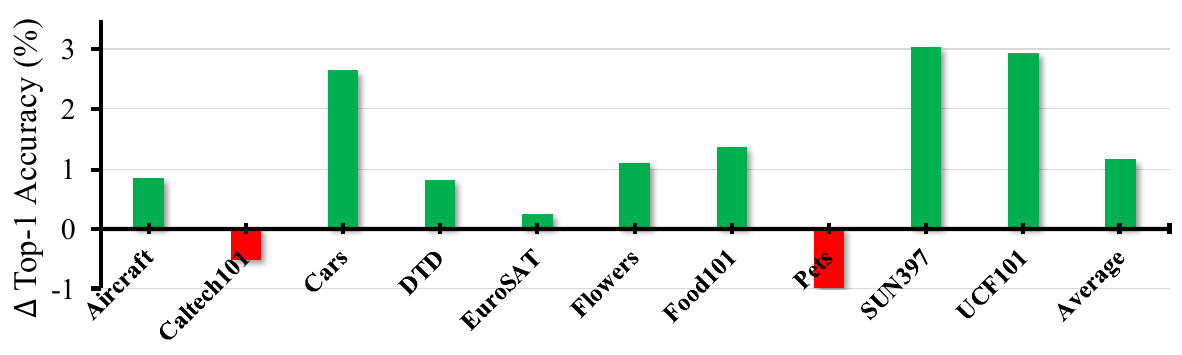}
    \caption{Impact of the calibration stage. Top-1 accuracy gains from adding calibration on top of exploration across fine-grained datasets.}
    \label{fig:ab_calib}
\end{figure}

\begin{table}[t]
\centering
\small
\caption{Ablation on loss components. Average top-1 accuracy (\%) on the fine-grained benchmarks.}
\label{tab:ablation_losses}
\resizebox{0.35\textwidth}{!}{
\begin{tabular}{l|c}
\toprule
Setting & Avg. Acc. (\%) \\
\midrule
\rowcolor{blue!8}
Full FCL (with $\mathcal{L}_{\text{cal}}$ + $\mathcal{L}_{\text{align}}$) & \textbf{65.83} \\
w/o Calibration Loss $\mathcal{L}_{\text{cal}}$ & 64.46 \\
w/o Alignment Loss $\mathcal{L}_{\text{align}}$ & 64.58 \\
\bottomrule
\end{tabular}}
\end{table}

We assess the contribution of our non–entropy-based calibration stage by comparing an \emph{exploration-only} variant—using two exploration passes to form $\mathcal{C}_K$ and predicting with $\Phi_1(\cdot)$—against the full \emph{exploration + calibration} pipeline. As shown in \autoref{fig:ab_calib}, eight of the ten fine-grained datasets exhibit accuracy gains (up to +3.05 pp), while the remaining two show only minor decreases ($<$1 pp), indicating that the calibration stage generally shifts performance in a positive direction. On average, calibration yields an improvement of +1.15 pp over exploration alone—accounting for roughly 54\% of the total gain over zero-shot CLIP (+2.12 pp). These results confirm that both exploration and calibration are essential contributors to the overall effectiveness of FCL. 
Further, \autoref{tab:ablation_losses} shows that removing either the calibration loss or the alignment loss reduces performance, confirming that both terms play complementary roles in stabilizing and guiding context adaptation.

\section{Conclusions}

We introduced \textbf{Fair Context Learning}, a principled episodic TTA framework that separates exploration from calibration and adapts textual contexts by equalizing class sensitivities to shared visual evidence. Our results demonstrate that robust test-time calibration of text embeddings is achievable \emph{without} entropy minimization, challenging the central assumption underlying most prior CLIP-based TTA methods. Instead, FCL leverages multi-view consensus and fairness-driven context updates to mitigate overconfident drift and reduce bias caused by partial feature obsession.
\
While our method relies on an approximate additive evidence decomposition and estimated evidence maps—which may deviate under strong contextual interactions or non-additive visual cues—empirical results show that these approximations are sufficiently reliable to support effective adaptation. Developing more precise approximations of evidence decomposition remains an open challenge.


\footnotesize
\section*{Acknowledgements}
This work was supported in part by the DARPA Young Faculty Award, the National Science Foundation (NSF) under Grants \#2127780, \#2319198, \#2321840, \#2312517, \#2431561, and \#2235472, the Semiconductor Research Corporation (SRC), the Office of Naval Research through the Young Investigator Program Award, and Grants \#N00014-21-1-2225 and \#N00014-22-1-2067, Army Research Office Grant \#W911NF2410360. Additionally, support was provided by the Air Force Office of Scientific Research under Award \#FA9550-22-1-0253, along with generous gifts from Xilinx and Cisco.

\normalsize
\bibliography{references}

\clearpage
\setcounter{page}{1}
\section*{Appendix}

\section{Jensen--Shannon Divergence in Calibration}
\label{sec:js_appendix}

Our calibration objective employs the Jensen--Shannon divergence (JSD) to encourage balanced predictions on the common evidence used during pairwise calibration.  
Given two probability distributions $p$ and $q$ over the same label space, the JSD is defined as
\[
\mathrm{JS}(p \,\|\, q)
= \tfrac{1}{2}\,\mathrm{KL}\!\left(p \,\middle\|\, m\right)
+ \tfrac{1}{2}\,\mathrm{KL}\!\left(q \,\middle\|\, m\right),
\qquad
m = \tfrac{1}{2}(p + q),
\]
where $\mathrm{KL}(\cdot\|\cdot)$ denotes the Kullback--Leibler divergence.  
JSD is symmetric, bounded ($0 \le \mathrm{JS} \le \log 2$), and remains well-defined even when either distribution assigns zero probability to some outcomes. These properties make it suitable for stable optimization in our setting, where stochastic masking and augmented evidence maps can produce highly skewed or sparse probability assignments.

\section{Lower Bound on Class Probability from the Softmax Margin}
\label{sec:appendix_lower_bound}

This section derives the lower bound used in the main text relating the softmax class probability to the score margin between a class and its strongest competitor.

For an augmented view $u_k$ and class scores $s_{k,1}(\delta),\dots,s_{k,C}(\delta)$, the softmax probability of class $c$ is
\[
\pi(c \mid u_k,\delta)
= 
\frac{\exp\!\left(s_{k,c}(\delta)\right)}{\sum_{j=1}^C \exp\!\left(s_{k,j}(\delta)\right)}.
\]
Define the margin between class $c$ and its closest competitor as
\[
m_k(c;\delta)
=
s_{k,c}(\delta)
-
\max_{j \neq c} s_{k,j}(\delta).
\]

Using this margin, a standard inequality for the softmax yields a probability lower bound.  
Let $j^\star = \arg\max_{j\neq c} s_{k,j}(\delta)$ denote the nearest competing class.  
Then
\[
\sum_{j=1}^C \exp\!\left(s_{k,j}(\delta)\right)
\le 
\exp\!\left(s_{k,c}(\delta)\right)
+
(C-1)\exp\!\left(s_{k,j^\star}(\delta)\right).
\]
Substituting into the softmax expression gives
\[
\pi(c \mid u_k,\delta)
\ge
\frac{\exp(s_{k,c}(\delta))}
{\exp(s_{k,c}(\delta)) + (C-1)\exp(s_{k,j^\star}(\delta))}.
\]
Factoring out $\exp(s_{k,c}(\delta))$ yields the closed-form lower bound
\[
\pi(c \mid u_k,\delta)
\ge
\frac{1}{1 + (C-1)\exp\!\left(-m_k(c;\delta)\right)}.
\]

This bound shows that the softmax confidence is controlled from below by the score margin: larger margins enforce higher minimum probability for class $c$.  
In the context of our analysis, this provides a direct link between low-entropy (high-confidence) augmented views and the presence of strong, discriminative class evidence in those views.

\section{Dataset Details}

\autoref{tab:dataset_stats} summarizes all datasets used in our evaluation.  
For ImageNet, we use the official validation set along with four widely adopted natural-distribution variants (ImageNet-A, ImageNet-Sketch, ImageNet-R, and ImageNet-V2).  
For all fine-grained datasets, we follow the standardized test splits provided by Zhou \emph{et al.}~\cite{zhou2022learning}.

The datasets span a broad range of recognition challenges, including fine-grained object categories (Aircraft, Cars, Pets), visually diverse real-world classes (Food101, SUN397), textures (DTD), satellite imagery (EuroSAT), and action-centric frames (UCF101).  
The ImageNet variants provide complementary natural distribution shifts, covering adversarial natural examples (ImageNet-A), artistic renditions (ImageNet-R), updated test samples (ImageNet-V2), and sketch-style drawings (ImageNet-Sketch).

\begin{table}[t]
\centering
\small
\caption{Dataset statistics for ImageNet variants and fine-grained benchmarks.}
\label{tab:dataset_stats}
\begin{tabular}{l|c|c}
\toprule
\textbf{Dataset} & \textbf{\# Classes} & \textbf{\# Eval Samples} \\
\midrule
ImageNet~\cite{deng2009imagenet} & 1{,}000 & 50{,}000 \\
ImageNet-A~\cite{hendrycks2021natural} & 200 & 7{,}500 \\
ImageNet-Sketch~\cite{wang2019learning} & 1{,}000 & 50{,}889 \\
ImageNet-R~\cite{hendrycks2021many} & 200 & 30{,}000 \\
ImageNet-V2~\cite{recht2019imagenet} & 1{,}000 & 10{,}000 \\
\midrule
FGVC-Aircraft~\cite{maji2013fine} & 100 & 3{,}333 \\
Caltech101~\cite{fei2004learning} & 100 & 2{,}465 \\
Stanford Cars~\cite{krause20133d} & 196 & 8{,}041 \\
DTD~\cite{cimpoi2014describing} & 47 & 1{,}880 \\
EuroSAT~\cite{helber2019eurosat} & 10 & 5{,}400 \\
Oxford Flowers~\cite{nilsback2008automated} & 102 & 2{,}463 \\
Food101~\cite{bossard2014food} & 101 & 25{,}250 \\
Oxford Pets~\cite{parkhi2012cats} & 37 & 3{,}680 \\
SUN397~\cite{xiao2010sun} & 397 & 19{,}850 \\
UCF101~\cite{soomro2012ucf101} & 101 & 3{,}783 \\
\bottomrule
\end{tabular}
\end{table}

\section{Baseline Details}

\begin{table}[t]
\centering
\small
\caption{Baseline methods used in our evaluation and their code availability.}
\label{tab:baseline_code}
\begin{tabular}{l|c|c}
\toprule
\textbf{Method} & \textbf{Venue} & \textbf{Code Available} \\
\midrule
TPT~\cite{shu2022test} & NeurIPS'22 & \checkmark \\
C{-}TPT~\cite{yoon2024c} & ICLR'24 & \checkmark \\
RLCF~\cite{zhao2024testtime} & ICLR'24 & \checkmark \\
MTA~\cite{zanella2024test} & CVPR'24 & \checkmark \\
ZERO~\cite{farina2024frustratingly} & NeurIPS'24 & \checkmark \\
TTL~\cite{imam2025test} & WACV'25 & \checkmark \\
TPS~\cite{sui2025just} & WACV'25 & \checkmark \\
R{-}TPT~\cite{sheng2025r} & CVPR'25 & \checkmark \\
\bottomrule
\end{tabular}
\end{table}

\begin{table}[t]
\centering
\footnotesize
\caption{Validation of Additive Evidence Decomposition (ImageNet val, averaged over 1k samples).}\label{tab:decomposition}
\resizebox{\columnwidth}{!}{
\begin{tabular}{l|c}
\toprule
Component & Cosine Sim. to $z$ \\
\midrule
Common evidence proxy ($\tilde z^{\mathrm{com}}$) & $0.7283$ \\
Unique proxy (top-1) ($\tilde z^{\mathrm{uniq}}_1$) & $0.6390$ \\
Unique proxy (top-2) ($\tilde z^{\mathrm{uniq}}_2$) & $0.6516$ \\
\midrule
\rowcolor{blue!8}
\textbf{Proxy sum (reconstruction)} ($\tilde{z}^{\text{com}} + 0.1 \tilde{z}^{\text{uniq}}_{u1} + 0.1 \tilde{z}^{\text{uniq}}_{u2}$) & $\mathbf{0.8705}$ \\
\bottomrule
\end{tabular}}
\end{table}

\begin{table*}[t]
\centering
\footnotesize
\caption{Computational Cost Analysis (ImageNet val, avg over 30 trials, NVIDIA RTX A6000).}
\label{tab:efficiency}
\resizebox{0.8\textwidth}{!}{
    \begin{tabular}{l|c|ccc|ccc}
        \toprule
        \multirow{2}{*}{\textbf{Method}} & \multirow{2}{*}{\textbf{Setting}} & \multicolumn{3}{c|}{\textbf{ViT-B/16}} & \multicolumn{3}{c}{\textbf{ViT-B/32}} \\
         & & \textbf{Time (s)} & \textbf{Peak Mem (GB)} & \textbf{Acc (\%)} & \textbf{Time (s)} & \textbf{Peak Mem (GB)} & \textbf{Acc (\%)} \\
        \midrule
        TPT~[31] & 1 step & $0.78 \pm 0.0045$ & $12.96$ & 68.94 & $0.77 \pm 0.0042$ & $12.97$ & 63.47 \\
        \midrule
        \rowcolor{blue!8}
        \textbf{Ours (FCL)} & $N_m=200$ & $\mathbf{0.46} \pm 0.0054$ & $\mathbf{1.05}$ & 69.43 & $\mathbf{0.24} \pm 0.0034$ & $\mathbf{0.87}$ & 65.50 \\
        \rowcolor{blue!8}
        \textbf{Ours (FCL)} & $N_m=400$ & $0.64 \pm 0.0022$ & $\mathbf{1.05}$ & \textbf{69.46} & $0.31 \pm 0.0036$ & $\mathbf{0.87}$ & \textbf{65.55} \\
        \rowcolor{blue!8}
        \textbf{Ours (FCL)} & $N_m=800$ & $1.04 \pm 0.0037$ & $\mathbf{1.05}$ & \textbf{69.47} & $0.45 \pm 0.0036$ & $\mathbf{0.87}$ & \textbf{65.55} \\
        \bottomrule
    \end{tabular}}
\end{table*}

\begin{table*}[t]
\centering
\caption{Top-1 accuracy (\%) of test-time adaptation methods on ImageNet and its distribution shift variants with CLIP-ViT-B/32. 
The best result for each dataset is shown in \textbf{bold}, and the second best is \underline{underlined}. 
OOD Avg.\ is the mean over ImageNet-A, -K, -R, and -V.}
\label{tab:tta_imagenet_rn50}
\resizebox{0.8\textwidth}{!}{
\begin{tabular}{l|ccccc|cc}
\toprule
Method & ImageNet & ImageNet-A & ImageNet-K & ImageNet-R & ImageNet-V & Avg. & OOD Avg. \\
\midrule
\rowcolor{gray!15}
CLIP~\cite{radford2021learning} & 62.04 & 29.53 & 40.84 & 66.23 & 54.77 & 50.68 & 47.84 \\
\midrule
TPT~\cite{shu2022test} & 63.47 & 34.68 & 41.67 & 69.01 & 56.85 & 53.14 & 50.55 \\
C-TPT~\cite{yoon2024c} & 63.85 & 32.43 & 42.18 & 68.45 & 56.42 & 52.67 & 49.87 \\
RLCF~\cite{zhao2024testtime} & 63.64 & 37.49 & 42.75 & 70.45 & 57.38 & 54.34 & 52.02 \\
MTA~\cite{zanella2024test} & 65.02 & 38.03 & 43.45 & 70.37 & 58.21 & 55.02 & 52.52 \\
ZERO~\cite{farina2024frustratingly} & \underline{65.29} & \underline{40.48} & \underline{43.65} & 70.76 & \underline{58.91} & \underline{55.82} & \underline{53.45} \\
TTL~\cite{imam2025test} & 64.99 & 39.65 & \textbf{43.84} & \underline{71.16} & \underline{58.91} & 55.71 & 53.39 \\
TPS~\cite{sui2025just} & 64.77 & 38.61 & 43.14 & 70.21 & 57.75 & 54.90 & 52.43 \\
R-TPT~\cite{sheng2025r} & 64.25 & 36.61 & 41.60 & 69.94 & 57.96 & 54.07 & 51.53 \\
\midrule
\rowcolor{blue!8}
Ours & \textbf{65.55} & \textbf{42.66} & 43.49 & \textbf{71.40} & \textbf{59.01} & \textbf{56.42} & \textbf{54.14} \\
\bottomrule
\end{tabular}}
\end{table*}

\begin{table}[t]
\centering
\small
\caption{Average and OOD-average accuracy deltas (\%) over zero-shot CLIP for two CLIP backbones.}
\label{tab:delta_clip_merged}
\begin{tabular}{l|cc|cc}
\toprule
\multirow{2}{*}{Method} & \multicolumn{2}{c|}{Avg.\ $\Delta$} & \multicolumn{2}{c}{OOD Avg.\ $\Delta$} \\
 & ViT-B/16 & ViT-B/32 & ViT-B/16 & ViT-B/32 \\
\midrule
TPT & +3.32 & +2.46 & +3.60 & +2.71 \\
C-TPT & +2.06 & +1.99 & +2.13 & +2.03 \\
RLCF & +3.71 & +3.66 & +4.17 & +4.18 \\
MTA & +3.97 & +4.34 & +4.33 & +4.68 \\
ZERO & +4.74 & +5.14 & +5.28 & +5.61 \\
TTL & +4.73 & +5.03 & +5.27 & +5.55 \\
TPS & +4.08 & +4.22 & +4.57 & +4.59 \\
R-TPT & +4.05 & +3.39 & +4.40 & +3.69 \\
\midrule
\rowcolor{blue!8}
Ours (FCL) & \textbf{+5.12} & \textbf{+5.74} & \textbf{+5.71} & \textbf{+6.30} \\
\bottomrule
\end{tabular}
\end{table}

\begin{table*}[t]
\centering
\caption{Top-1 accuracy (\%) of test-time adaptation methods on ImageNet and its distribution-shift variants. 
The best result for each dataset is in \textbf{bold}; the second best is \underline{underlined}. 
OOD Avg.\ is the mean over ImageNet-A, -K, -R, and -V.}
\label{tab:tta_context_hardprompt}
\resizebox{0.82\textwidth}{!}{
\begin{tabular}{l|ccccc|cc}
\toprule
Method & ImageNet & ImageNet-A & ImageNet-K & ImageNet-R & ImageNet-V & Avg. & OOD Avg. \\
\midrule
\rowcolor{gray!15}
CLIP~\cite{radford2021learning} & 66.72 & 47.80 & 46.15 & 73.99 & 60.84 & 59.10 & 57.20 \\
\midrule
TPT~\cite{shu2022test} & 68.94 & 54.75 & 47.92 & 77.13 & 63.38 & 62.42 & 60.80 \\
C-TPT~\cite{yoon2024c} & 68.52 & 51.35 & 47.48 & 75.87 & 62.60 & 61.16 & 59.33 \\
RLCF~\cite{zhao2024testtime} & 68.56 & 57.39 & 47.98 & 77.08 & 63.02 & 62.81 & 61.37 \\
MTA~\cite{zanella2024test} & 69.24 & 57.03 & 48.48 & 77.00 & 63.60 & 63.07 & 61.53 \\
ZERO~\cite{farina2024frustratingly} & 69.28 & 59.77 & 48.59 & 77.38 & 64.16 & 63.84 & 62.48 \\
TTL~\cite{imam2025test} & 69.28 & 59.00 & \underline{48.75} & 77.76 & \underline{64.36} & 63.83 & 62.47 \\
TPS~\cite{sui2025just} & 68.83 & 58.19 & 48.24 & 76.98 & 63.68 & 63.18 & 61.77 \\
R-TPT~\cite{sheng2025r} & 69.37 & 57.72 & 47.75 & 76.93 & 63.98 & 63.15 & 61.60 \\
\midrule
\rowcolor{blue!8}
Ours (CL) & \underline{69.46} & \underline{61.38} & 48.25 & \underline{77.78} & 64.23 & \underline{64.22} & \underline{62.91} \\
\rowcolor{blue!16}
Ours (CL-HP) & \textbf{69.67} & \textbf{62.50} & \textbf{48.81} & \textbf{78.99} & \textbf{64.72} & \textbf{64.93} & \textbf{63.75} \\
\bottomrule
\end{tabular}}
\end{table*}

\begin{figure*}[t]
    \centering
    \includegraphics[width=1.0\textwidth]{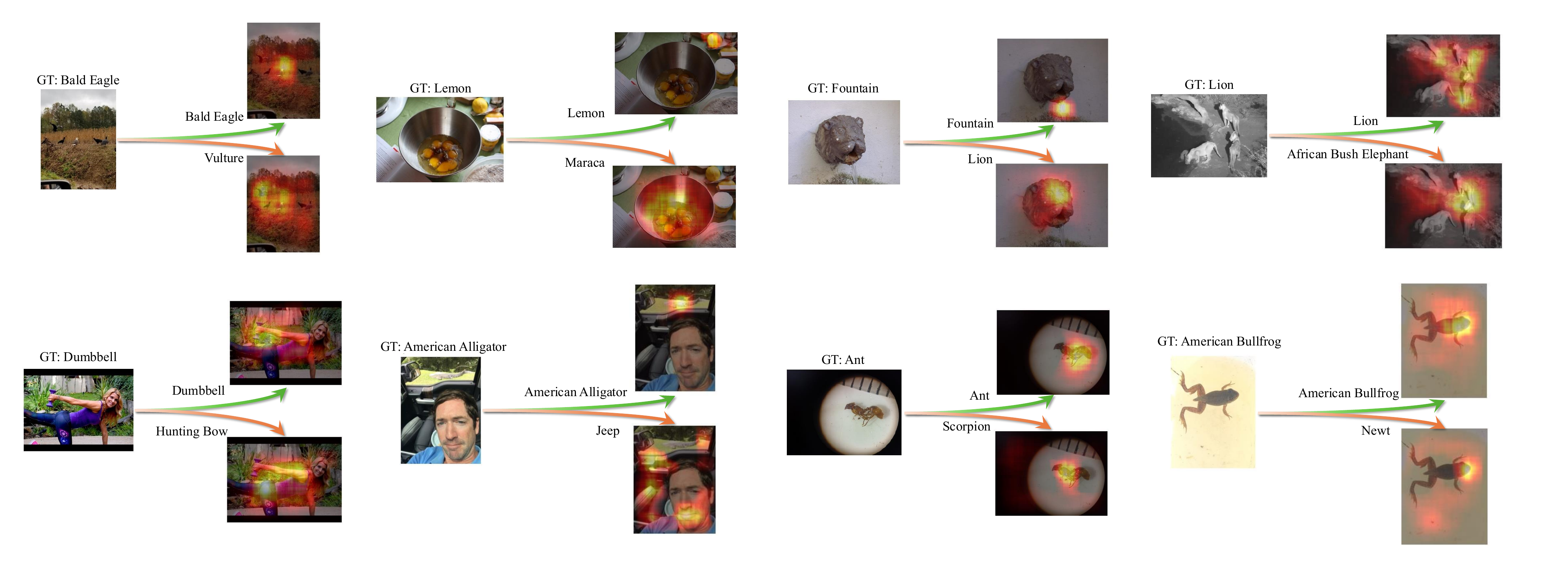}
    \caption{Visualization of estimated class evidence maps for randomly sampled ImageNet-A images.
For each example, we show the original image (left), the ground-truth class evidence map (top right), and the evidence map of a competing incorrect class (bottom right).
Brighter regions indicate areas the model relies on more heavily when making its prediction.}
    \label{fig:fcl_evidencemaps}
\end{figure*}

\begin{table}[t]
\centering
\caption{Effect of aggregation strategy during exploration on ImageNet accuracy (\%).}
\label{tab:ab_aggregation}
\small
\begin{tabular}{l|c}
\toprule
Method & ImageNet \\
\midrule
\rowcolor{gray!15}
CLIP~\cite{radford2021learning} & 66.72 \\
\midrule
Ours (Mean Aggregation) & 67.36 \\
\rowcolor{blue!8}
Ours (Voting Aggregation) & \textbf{69.46} \\
\bottomrule
\end{tabular}
\end{table}

\begin{table}[t]
\centering
\caption{Effect of the number of random masks $N_m$ for class evidence estimation on ImageNet accuracy (\%).}
\label{tab:ab_masks}
\small
\begin{tabular}{l|c}
\toprule
Setting & ImageNet \\
\midrule
\rowcolor{gray!15}
CLIP~\cite{radford2021learning} & 66.72 \\
\midrule
Ours with $N_m=50$ & 68.13 \\
Ours with $N_m=100$ & 69.02 \\
\rowcolor{blue!4}
Ours with $N_m=200$ & 69.43 \\
\rowcolor{blue!8}
Ours with $N_m=400$ & \textbf{69.46} \\
\rowcolor{blue!16}
Ours with $N_m=800$ & \textbf{69.47} \\
\bottomrule
\end{tabular}
\end{table}

\begin{table}[t]
\centering
\caption{ImageNet top-1 accuracy (\%) with a 7-prompt ensemble built from generic CLIP templates.}
\label{tab:ensemble}
\small
\begin{tabular}{l|c}
\toprule
Method & ImageNet \\
\midrule
\rowcolor{gray!15}
CLIP~\cite{radford2021learning} & 66.72 \\
\midrule
\rowcolor{blue!8}
Ours (w/o Ensemble) & 69.46 \\
\rowcolor{blue!16}
Ours (CL + Ensemble) & 71.04 \\
\rowcolor{blue!24}
Ours (CL-HP + Ensemble) & \textbf{71.53} \\
\bottomrule
\end{tabular}
\end{table}

All baselines included in our evaluation are state-of-the-art TTA methods for VLMs with publicly available code, as shown in \autoref{tab:baseline_code}. 
Below, we briefly summarize each method and clarify how our approach differs.

\paragraph{TPT~\cite{shu2022test}.}
Test-Time Prompt Tuning optimizes the textual context by minimizing entropy across augmented views. While effective, its confidence-maximization objective can amplify biased shared evidence, reinforcing overconfident misclassification under partial feature obsession. FCL avoids entropy minimization entirely and instead calibrates text embeddings using shared-evidence–aware constraints.

\paragraph{C-TPT~\cite{yoon2024c}.}
This method adds a calibration regularizer to TPT, encouraging stability of predictions across augmentations. However, it still relies on entropy minimization as its primary signal. From our evidence-decomposition view, such confidence-centric updates remain sensitive to shared feature bias, whereas FCL explicitly equalizes sensitivities to common evidence.

\paragraph{RLCF~\cite{zhao2024testtime}.}
RLCF uses reinforcement learning guided by CLIP-based rewards to update prompts. Although it avoids direct entropy optimization, it incurs substantial computational overhead and depends on the teacher model's reliability. In contrast, FCL uses a lightweight analytic calibration objective grounded in estimated common-evidence maps.

\paragraph{MTA~\cite{zanella2024test}.}
MTA aggregates predictions from multiple augmentations via feature averaging to reduce variance. While efficient, it does not correct systematic biases caused by shared features, which our evidence formulation identifies as a core failure mode. FCL instead performs fairness-driven calibration to suppress such bias.

\paragraph{ZERO~\cite{farina2024frustratingly}.}
ZERO replaces optimization with majority voting over confident views, motivated by a Bernoulli independence assumption. Our analysis shows that this assumption breaks when augmented views emphasize correlated common evidence. FCL retains the robustness of voting for exploration but adds a calibration stage to address shared-feature bias.

\paragraph{TTL~\cite{imam2025test}.}
TTL performs lightweight test-time low-rank adaptation of model parameters. While efficient, it lacks mechanisms to disentangle common vs.\ unique evidence, making it susceptible to the same bias amplification observed in prompt-tuning methods. FCL targets this issue directly via fairness-based text adaptation.

\paragraph{TPS~\cite{sui2025just}.}
TPS shifts text embeddings using entropy-weighted feature aggregation. Its reliance on entropy still ties the optimization to confidence amplification. FCL avoids entropy altogether and instead constrains sensitivity to shared evidence across candidate classes.

\paragraph{R-TPT~\cite{sheng2025r}.}
R-TPT improves TPT by reducing entropy independently for each augmented view. However, its optimization remains confidence-driven. Under our theoretical lens, R-TPT cannot correct systematic bias arising from strong shared evidence. FCL explicitly mitigates this by enforcing evidence-balanced calibration.

\medskip
Overall, most baselines rely—directly or indirectly—on confidence maximization or aggregation assumptions that overlook correlations induced by shared visual features. FCL differs by grounding adaptation in an evidence-decomposition perspective and applying fairness-based calibration to counteract shared-evidence bias.

\section{Further Empirical Validation of Proxy Evidence Decomposition}
\label{sec:appendix_proxy_validation}

This section provides additional empirical support for the proxy-based additive evidence decomposition used in our calibration framework.
While the decomposition in \autoref{eq:additivedecomp} is not claimed to be exact, we show that the proposed proxy components capture the dominant structure of the image embedding and are sufficiently accurate for evidence-aware calibration.
\
We evaluate the quality of the proxy decomposition by measuring how well the estimated common- and unique-evidence embeddings reconstruct the original image embedding.
Using the proxy common evidence $\tilde z^{\mathrm{com}}$ and the top two unique-evidence proxies $\tilde z^{\mathrm{uniq}}_{1}, \tilde z^{\mathrm{uniq}}_{2}$, we form a simple linear reconstruction with fixed weights.
As reported in \autoref{tab:decomposition}, the reconstructed embedding achieves high cosine similarity to the original embedding, substantially exceeding that of any single component.
This indicates that the proxy components preserve complementary and meaningful information rather than arbitrary noise.

\section{Computational Efficiency and Overhead}
\label{sec:appendix_efficiency}

This section summarizes the computational characteristics of FCL and explains why the method remains efficient in practice.
\
The class-evidence estimation stage in FCL is entirely gradient-free and consists only of forward passes of the visual encoder on randomly masked inputs.
As a result, it incurs no backpropagation cost and introduces limited memory overhead.
In contrast, entropy-minimization–based test-time adaptation methods backpropagate through the text encoder over the full label space for every augmented view, which scales poorly with the number of classes.
\
FCL further improves efficiency by restricting all gradient-based optimization to a small candidate set selected during exploration.
With a fixed candidate size of $K{=}10$, the calibration stage reduces gradient computation by roughly two orders of magnitude compared to full-label-space prompt tuning.
This design ensures that the additional cost of calibration remains modest even when many augmented views are used.
\
Empirically, as shown in \autoref{tab:efficiency}, FCL is faster than TPT on ImageNet with 64 views and uses approximately $12\times$ less peak GPU memory across both ViT-B/16 and ViT-B/32 backbones, while achieving higher accuracy.
Increasing the number of random masks improves evidence estimation quality but does not increase peak memory usage, since masking-based estimation relies only on forward activations.
\
Taken together, these results show that FCL achieves evidence-aware calibration with favorable computational efficiency, combining a lightweight gradient-free evidence estimation stage with highly localized, candidate-restricted optimization.

\section{Generalization Across Model Architectures}

To assess whether our findings extend beyond a single architecture, we further evaluate FCL using CLIP ViT-B/32, a lower-capacity encoder with reduced spatial resolution and representational expressiveness. This setup provides a stricter test of whether adaptation gains persist when the underlying visual features are less informative.

As shown in \autoref{tab:tta_imagenet_rn50}, FCL consistently improves accuracy across all ImageNet variants even under this weaker backbone. FCL achieves the best performance on ImageNet, ImageNet-A, ImageNet-R, and ImageNet-V, and attains the highest overall and OOD averages. These results suggest that the benefits of evidence-balanced calibration are not dependent on high-resolution features or stronger base representations.

\autoref{tab:delta_clip_merged} reports the average and OOD-average improvements over the corresponding zero-shot CLIP baselines for both ViT-B/16 and ViT-B/32.
Across both backbones, FCL yields the largest relative gains. This indicates that the effectiveness of our approach is driven by the mechanism of evidence-balanced calibration itself, rather than characteristics specific to a particular encoder architecture.

\section{Qualitative Visualization of Class Evidence Maps}
\label{sec:evidence_maps}

To examine the reliability of our class evidence estimation, we visualize the estimated evidence maps by overlaying them as transparent heatmaps on the input images (\autoref{fig:fcl_evidencemaps}).
Across diverse ImageNet-A samples, class evidence maps for mispredicted classes frequently highlight regions containing spurious or only partially correlated visual cues, aligning with our empirical observation that CLIP can rely on partial features under distribution shift.
In contrast, the ground-truth class maps generally highlight more semantically aligned object regions.
While these qualitative patterns do not constitute a definitive causal analysis, they provide empirical support that our estimation procedure captures meaningful aspects of the model's internal evidence distribution—an important prerequisite for the fairness-oriented calibration used in FCL.

\section{Context Learning with Hard Prompt (CL-HP)}

To study how different prompt parameterizations affect Fair Context Learning, we evaluate a variant that preserves the hard prompt (\emph{``a photo of a''}) while prepending a small number of learnable continuous context tokens. We refer to this approach as \textbf{Context Learning with Hard Prompt (CL-HP)}. Unlike prefix learning, which updates the embeddings of the hard prompt tokens themselves, CL-HP maintains the fixed semantic structure of the prompt while enabling adaptation through softly learned contextual tokens.

Using four learnable context tokens, CL-HP yields consistently stronger performance than both our earlier context-learning variant (CL) and all existing TTA baselines. As shown in \autoref{tab:tta_context_hardprompt}, CL-HP achieves the highest accuracy across ImageNet and all four distribution-shift variants. From the viewpoint of our fairness-driven calibration formulation, a plausible interpretation is that fixing the semantic prompt may help anchor the class representation, while the additional contextual tokens provide a controlled way to adjust sensitivity to shared evidence. This separation could reduce susceptibility to spurious feature drift during adaptation, offering a potential explanation for the more stable updates observed with CL-HP.

\section{Additional Ablations}
\label{sec:more_ablations}

This section provides supplementary ablations examining two components of our framework: (i) the aggregation strategy used during exploration and (ii) the number of random masks employed for class-evidence estimation. Results in \autoref{tab:ab_aggregation} and \autoref{tab:ab_masks} demonstrate the robustness of these design choices.

\paragraph{Aggregation Strategy.}
\autoref{tab:ab_aggregation} compares mean aggregation and majority voting for selecting high-confidence views.
Both approaches outperform the zero-shot CLIP baseline; however, voting yields a noticeably larger improvement (+2.74 pp over CLIP and +2.10 pp over mean).
This suggests that voting more effectively captures agreement across stochastic augmentations and is less susceptible to outlier views, consistent with our interpretation of exploration as a process for identifying stable candidate hypotheses.

\paragraph{Number of Random Masks.}
\autoref{tab:ab_masks} evaluates the effect of varying the number of random masks $N_m$ used to estimate class evidence maps.
Accuracy improves steadily as $N_m$ increases from 50 to 200, after which additional masks produce diminishing returns.
The difference between $N_m=400$ and $N_m=800$ is minimal ($<0.05$ pp), indicating that evidence estimation has largely converged.
These observations suggest that our choice of $N_m=400$ achieves a practical balance between estimation quality and computational cost.

\section{Prompt Ensemble with Generic Templates}
\label{sec:appendix_ensemble}

In all main experiments we use a single hard prompt (``a photo of a \{\textit{class}\}'') for constructing class descriptions.  
To assess whether FCL can benefit from textual diversity without relying on dataset-specific prompt engineering, we further evaluate a prompt ensemble using the seven generic templates provided in the official CLIP repository.
For each template, we build class prompts, apply FCL (either CL or CL-HP) independently, and obtain a per-template prediction.  
Then, we aggregate predictions across templates by majority vote over class labels.  
\autoref{tab:ensemble} reports ImageNet top-1 accuracy under this prompt ensemble.

Relative to zero-shot CLIP, the ensemble improves accuracy by +4.32 pp for CL and +4.81 pp for CL-HP.  
Compared to their single-prompt counterparts, the ensemble yields additional gains of roughly 1.5–2.0 pp.  
These results indicate that FCL can exploit diversity in generic textual contexts in a complementary way to its evidence-balanced calibration: multiple generic prompts provide varied linguistic views of the same class, and FCL's calibration over shared visual evidence appears compatible with aggregating these views at the prediction level, without dataset-specific prompt design.

\end{document}